\documentclass[conference]{IEEEtran}
\IEEEoverridecommandlockouts
\usepackage{cite}
\usepackage{amsmath,amssymb,amsfonts}
\usepackage{algorithmic}
\usepackage{algorithm}
\usepackage{graphicx}
\usepackage{textcomp}
\usepackage{xcolor}
\usepackage{comment}
\usepackage{balance}

\usepackage{fancyhdr}
\usepackage{ragged2e}

\def\BibTeX{{\rm B\kern-.05em{\sc i\kern-.025em b}\kern-.08em
    T\kern-.1667em\lower.7ex\hbox{E}\kern-.125emX}}
\begin{document}

\newcommand{\changed}[1]{\textcolor[rgb]{0.8,0.0,0.4}{{}{#1}}}
\newcommand{\todo}[1]{\textcolor[rgb]{0.0,0.8,0.4}{\small {[TODO]}{#1}}}

\fancypagestyle{firstpage}
{
    \fancyhf{}

    \fancyhead[L]{\small \centering \noindent Preprint accepted at the 41st International Conference of the Chilean Computer Science Society, SCCC 2022, Santiago, Chile, 2022. 
    }

}

\title{Explaining Agent's Decision-making in a Hierarchical Reinforcement Learning Scenario
}

\author{\IEEEauthorblockN{Hugo Muñoz\IEEEauthorrefmark{1}, Ernesto Portugal\IEEEauthorrefmark{2}, Angel Ayala\IEEEauthorrefmark{2}, Bruno Fernandes\IEEEauthorrefmark{2} and Francisco Cruz\IEEEauthorrefmark{3}\IEEEauthorrefmark{1}}
\IEEEauthorblockA{\IEEEauthorrefmark{1}Escuela de Ingenier\'ia, Universidad Central de Chile, Santiago, Chile\\
Email: hugo.munoz@alumnos.ucentral.cl}
\IEEEauthorblockA{\IEEEauthorrefmark{2}Escola Polit\'ecnica de Pernambuco, Universidade de Pernambuco, Recife, Brasil\\
Emails: \{eipb, aaam, bjtf\}@ecomp.poli.br}
\IEEEauthorblockA{\IEEEauthorrefmark{3}School of Computer Science and Engineering, University of New South Wales, Sydney, Australia\\
Email: f.cruz@unsw.edu.au}}

\maketitle
\IEEEpubidadjcol
\begin{abstract}
Reinforcement learning is a machine learning approach based on behavioral psychology.
It is focused on learning agents that can acquire knowledge and learn to carry out new tasks by interacting with the environment.
However, a problem occurs when reinforcement learning is used in critical contexts where the users of the system need to have more information and reliability for the actions executed by an agent. 
In this regard, explainable reinforcement learning seeks to provide to an agent in training with methods in order to explain its behavior in such a way that users with no experience in machine learning could understand the agent's behavior.
One of these is the memory-based explainable reinforcement learning method that is used to compute probabilities of success for each state-action pair using an episodic memory. 
In this work, we propose to make use of the memory-based explainable reinforcement learning method in a hierarchical environment composed of sub-tasks that need to be first addressed to solve a more complex task. 
The end goal is to verify if it is possible to provide to the agent the ability to explain its actions in the global task as well as in the sub-tasks. 
The results obtained showed that it is possible to use the memory-based method in hierarchical environments with high-level tasks and compute the probabilities of success to be used as a basis for explaining the agent's behavior. 
%\todo{Maybe we need to change the title, something like hierarchical environment and explaining or explainability.... Maybe explaining agent's decision-making in a hierarchical reinforcement learning scenario.} ... Hierarchical Reinforcement Learning: Justifying Decision Making by an Autonomous Agent.
\end{abstract}

%\begin{IEEEkeywords}
%component, formatting, style, styling, insert
%\end{IEEEkeywords}

\thispagestyle{firstpage}
\section{Introduction}
During the last decade, the growth of machine learning algorithms has been relevant to the point of being successful in areas of science as well as in the daily life of humans. 
Some areas include health, image processing, games, autonomous cars, recommended content, among others~\cite{dovsilovic2018explainable}. 
Many of these algorithms use artificial neural networks (ANN) that are known as a black box system~\cite{olden2002illuminating} or a combination of ANN and phenomenological models known as gray box systems~\cite{cruz2007indirect, naranjo2010indirect}.
An ANN allows to infer an output value that depends on a combination of input values, however, the process for determining the inference value is based on empirical data. 
As a result, these systems lack credibility in environments where the decisions need to be selected in such a way so that they are correct and grounded. 
For example, in the case of diagnosing illnesses, an algorithm needs to be capable of considering different alternatives and determine and catalog the test correctly.

Furthermore, there exist other autonomous learning algorithms mainly used for solving sequential decisions problems, known as reinforcement learning (RL)~\cite{sutton2018reinforcement}. 
In this type of algorithm, an agent needs to learn to make a decision through trial and error to solve a task formulated as a Markovian decision problem.
Thus, commencing from a state $s_{t}$, an agent chooses an action $a_t$ that allows the transition to the next state $s_{t+1}$, obtaining a reward signal $r_{t+1}$ to evaluate the quality of the action selected from state $s_{t}$. 
% The central idea is that by selecting  a determined state. 
The main idea is that through observation of the reward at each step, the agent will be capable of refining a policy that allows it to select actions to receive a higher reward at the end of the task~\cite{cruz2018action}. 
Similarly for the RL algorithms, a person without knowledge of  artificial intelligence does not know the form or aspects that are considered by the agent to select an action~\cite{barros2020moody}.
In RL methods, the hierarchical approach is based on the ability of cognitive beings to resolve complex challenges by dividing them into more tractable smaller parts.
In addition, it is possible to learn new tasks quickly through the sequence of the behaviors learned, although the task requires various low-level actions~\cite{frans2017meta}. 
For example, humans can learn new tasks quickly by classifying the parts learned, including even if the task requires millions of low-level actions, such as muscular contractions. 
Hierarchical reinforcement learning (HRL), an extension of RL, models these problems in order to make the agents represent complicated behaviors as a short sequence of high-level actions. 
As a result, the agent can solve more complex problems. 
Therefore, if some solutions require a great number of low-level actions, the hierarchical policy could be converted into a sequence of high-level actions~\cite{frans2017meta}.

In this regard, explainable artificial intelligence (XAI) is the area that seeks to provide the ability to those systems in order to be able to explain its behavior in such a way that it is understandable to humans~\cite{gunning2019xai, dazeley2021levels}. 
Likewise, explainable reinforcement learning (XRL) emerges as a sub-task of XAI~\cite{dazeley2021explainable}. 
Since this subarea is focused on RL, the methods for making the system capable of providing an explanation are based on stages from the learning process. 
These methods may be based on: relevant features, the learning and  Markov decision process (MDP), or at policy level~\cite{milani2022survey}. 

Diverse techniques have been used in order to be able to explain behavior in hierarchical environments. 
For instance, just as dividing the tasks into different levels where the highest level groups the lowest level tasks and trains an agent for each level~\cite{heuillet2021explainability}. 
Also, based on human behavior to navigate through a room or simply that the model learns to carry our basic tasks and afterwards put together these basic tasks to carry out new ones. 
In this research, we sought to provide another alternative so that the models would be capable of providing an explanation in hierarchical contexts. 
Using the memory-based explainable reinforcement learning method~\cite{cruz2019memory}, we proposed analyzing the probabilities of success for different low-level tasks in a hierarchy, in addition to obtaining a global probability of success, for the completed task. 
The probability of success may be used as a basis for explaining the behavior of an autonomous agent in a hierarchical environment. 
The explanations generated are offered in a natural language representation with the ability to be better understood by not only RL practitioners, but also by any person with no knowledge of the area~\cite{cruz2021explainable}.

Our main contribution is the extension of a memory-based explainability method for hierarchical scenarios. 
% that was previously tested on a bounded and unbounded grid world. Achieving a way to endow an agent to explain himself for his actions. 
The rest of the paper is structured as follows. 
The next section reviews the background and related works. 
The third section introduces the explainability method used in this work, i.e., hierarchical explainable reinforcement learning. 
Section 4 describes the experimental scenario and Section 5 shows the results obtained during experiments. 
Finally, Section 6 depicts the main conclusions and possible future work.

\section{Related works}
Previous works have studied explainability in RL algorithms using explanations in distinct levels, depending on the target audience and the type of task involved~\cite{dazeley2021explainable, cruz2022evaluating}. 
A recent study classified the XRL methods as transparent algorithms and post-hoc explainability. 
In the first case, the algorithms present a transparent architecture that allows them to explain directly from the method without the need of an external process.  
In the second case, the algorithms must be analyzed after the execution of the method. 
Another study~\cite{cruz2021explainable} was more focused on the explanation of responsibilities for goal-driven tasks based on a generic concept for different users in a robotic scenario. 
In this approach, a probability of success is computed for a robot action from a particular state to indicate the confidence of reaching a final state. 
These goal-driven explanations are separated as follows:

\begin{itemize}
    \item Introspection-based~\cite{ayala2021explainable}: where the probability of success is estimated directly from the Q-values obtained.
    \item Learning-based~\cite{portugal2022analysis}: where the probability of success is learned during the training process. 
    \item Memory-based~\cite{cruz2019memory}: where the probability of success is computed using the total number of transitions and the total number of transition within a correct sequence.
\end{itemize}

For more complex tasks, studies that have been published have used hierarchical reinforcement learning (HRL). 
HRL allows reducing the complexity of the task by dividing it into sub-tasks to solve the problem. 
In~\cite{zakershahrak2020we}, they assert that the process of understanding a complex explanation is hierarchical.
They use the concept of mentality to provide an explanation to a human through hierarchical XRL. 
They carried out experiments set in the scavenger-hunt domain by looking for a treasure where the robot looks for explanations at intentional and action levels. 
This study demonstrated that humans prefer explanations with different levels of detail. 

In the study~\cite{beyret2019dot}, hierarchical XRL was used as the basis to explain the decision-making function. 
They used Deep Deterministic Policy Gradient (DDPG). 
DDPG employs an ANN to predict two hierarchical levels (high level and low level), the sub-tasks and the goal. 
However, for researchers, the explainability level for humans is still considered very premature. 
Moreover, in~\cite{rietz2022hierarchical} is proposed a method to train an RL agent to give local explanations by deconstructing hierarchical goals. 
In their research, they used HRL to help the reward decomposition algorithm (drQ) to give explanations. 
Utilizing two hierarchical levels, they employed a high-level agent to learn the sequences of the challenges in order to solve the task while the low-level agents were in charge of solving the challenges.

In summary, in the field of XRL, we can classify the methods into two groups: transparent algorithm or post-hoc explainability. 
Nonetheless, although a few methods have addressed providing explanations of RL algorithms, they differ from each other in terms of what they want to explain and to whom. 
Particularly in the field of HRL, the general view of explainability is still missing. 
Our approach extends one of those XRL methods to a hierarchical context.

\section{Hierarchical explainable reinforcement learning}

Our proposed method of explainability was implemented for an hierarchical reinforcement learning algorithm (HRL). 
The basis for an RL algorithm is that an agent needs to find ways to solve a problem through trial and error by maximizing the reward obtained by executing a determined action. 
Thus, the problem is formulated as a Markov decision process where the agent needs to find a policy $\pi$ that allows it to select an action $a_t$ on state $s_t$ to maximize the reward value obtained $r_{t+1}$. 
A Q-value function allows estimating the reward that can be obtained in the long run for each action $a$ given a state $s$, such as is explained in Equation~\eqref{Eq:accion-valor}.

\begin{equation}
    Q^{*}(s,a)=\underset{\pi }{max} \mathbb{E}[r_{t}+\gamma r_{t+1}+\gamma^{2} r_{t+2}+...|s_t=s,a_t=a,\pi],
    \label{Eq:accion-valor}
\end{equation}
where $\gamma \in [0,1]$ is the discount factor, a rate that sets how much the future rewards are considered.

With regard to the hierarchical method, we define low- and high-level actions.
In our scenario, the low-level actions correspond to the directions that the agent can move. 
Therefore, an estimation of the Q-value was defined for each one of these four actions executed. 
The high-level actions correspond to the different tasks that the agent should solve. 
These are three high-level tasks in the scenario proposed. 
Along with HRL method, the memory-based XRL method is used to provide explanations about the actions carried out by the apprentice agent in a determined state. 
Although the Q-values could be used to explain the behavior of the RL agent to the expert users in the area, our method looks for an agent capable of providing an explanation that makes sense to all types of users whether or not they know RL.
The explanations use the probability of successfully completing the task by selecting an action in a state, in addition to the number of transitions the agent carries out to accomplish the task. 
Once the task in finished, the agent may give an explanation based on probabilities of success; or a counterfactual explanation of why the agent selected one action over another, using more comprehensible language to a non-expert user. 

We proposed an memory-based explainable reinforcement learning algorithm in order to compute the probability of success using an RL agent with episodic memory. 
When accessing the agent's memory, its behavior may be understood as a result of its experience. 
For this a list of state-action pairs $T_{list}$ is implemented including all of the transitions the agent carried out during its learning process. 
Moreover, in order to compute the probability of success, the total number of transitions $T_t$ the agent made should be saved and the number of transitions in a sequences of success $T_s$. 
$T_t$ and $T_s$ correspond to a matrix of state-action pairs. 
Each time the agent reaches the final state, the probability of success $P_s$ is computed by dividing $T_s$ in $T_t$, i. e., $P_s \leftarrow T_s / T_t$~\cite{cruz2019memory}.

\begin{figure}%[htbp]
\centering
\includegraphics[width=\linewidth]{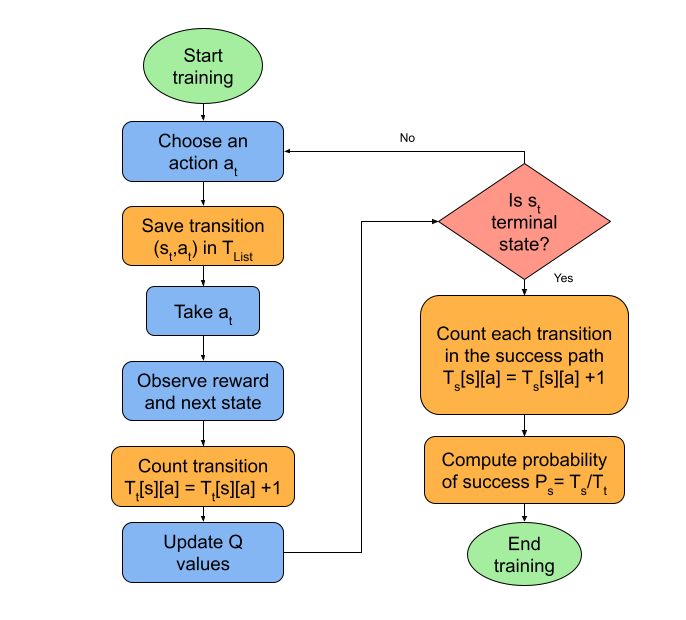}
\caption{Flowchart of the memory-based explainable reinforcement learning method. Blue boxes are part of the Q-learning algorithm and the orange boxes are the steps to implement the explainability method.}
\label{fig:diagrama}
\end{figure}

\begin{figure}%[htbp]
\centering
\includegraphics[width=\linewidth]{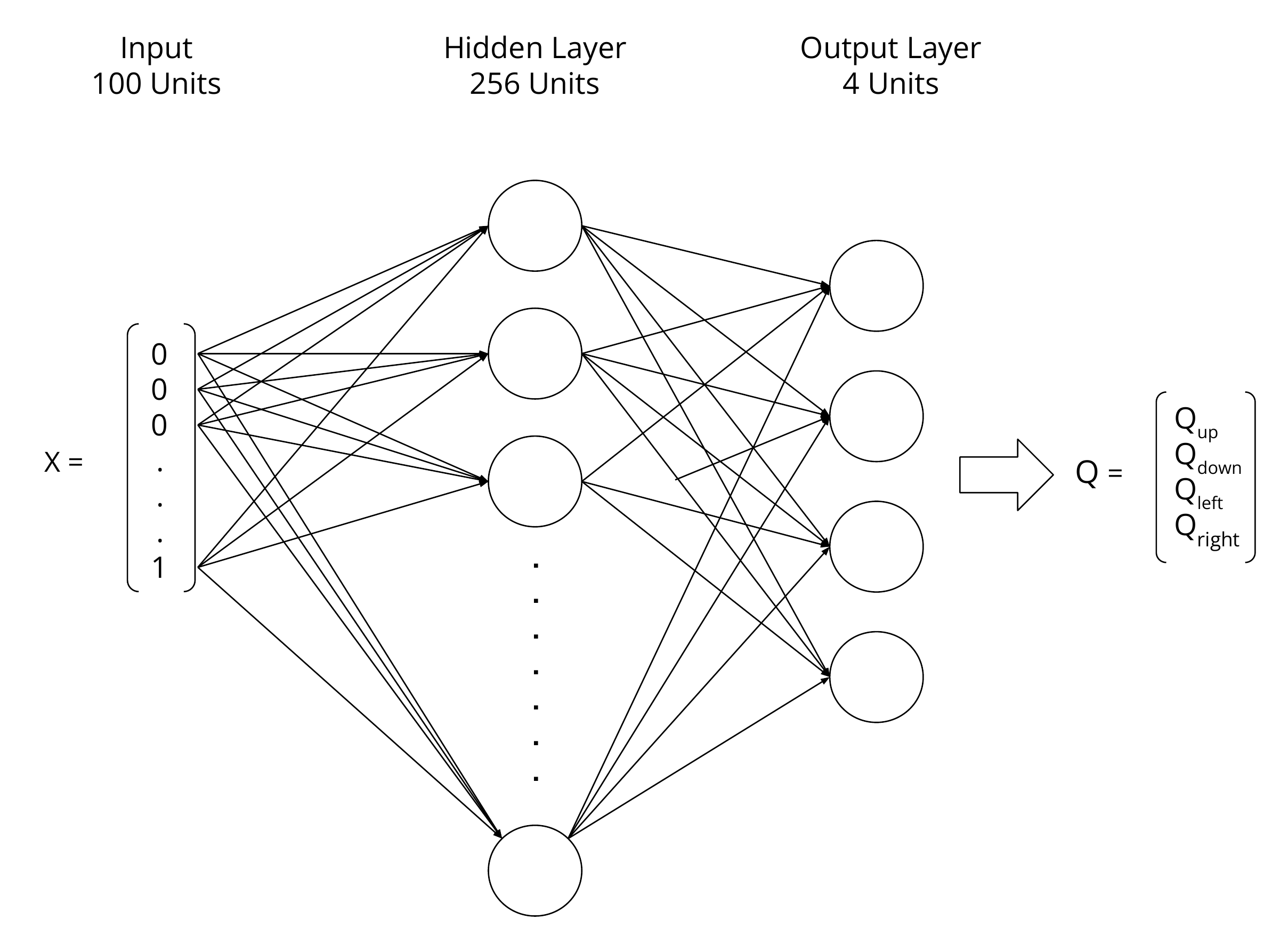}
\caption{Artificial neural network to compute the Q-values for four possible low-level actions. 
The network uses state $s_t$ as input setting to one the agent's position and zero otherwise. 
Subsequently, the input was processed with 256 neurons in the hidden layer, and obtaining four Q-values indicating the maximum future reward that may be obtained with each of them.}
\label{fig:modelo}
\end{figure}

The explainability method is used with the Q-learning algorithm~\cite{watkins1992q}.
The method selects an action for state $s_t$ using the tuple $<s_t, a_t, r_{t+1}, s_{t+1}>$. 
Algorithm~\ref{alg:learning} shows the memory-based explainable reinforcement learning method and Figure~\ref{fig:diagrama} illustrates more details about it.
The Q-values for each available action is estimated by a function approximation based on a neural network with 100 inputs, representing 100 possible states, 256 neurons in the hidden layer, and four outputs representing the Q-values for each possible low-level action. 
Figure~\ref{fig:modelo} shows the architecture of the network. 

\begin{algorithm} % enter the algorithm environment
\caption{Memory-based explainable reinforcement learning method~\cite{cruz2019memory}.} % give the algorithm a caption
\label{alg:learning} % and a label for~\ref{} commands later in the document
\begin{algorithmic}[1] % enter the algorithmic environment
    \STATE Initialize $Q(s,a), T_t, T_s, P_s$
    \FOR{each episode}
        \STATE Initialize $T_{list}$
        % \STATE \text{Choose an action using $a \leftarrow$ selectAction($s_t$)}
        \STATE Initialize $s_t \leftarrow s_0$
        \REPEAT
            % \STATE Choose an action using $a_t \leftarrow$ selectAction($s_t$)
            \STATE Choose an action $a_t$ from $s_t$ using $\epsilon$-greedy policy
            \STATE Take action $a_t$
            \STATE Save state-action transition in $T_{list}$
            \STATE Observe reward $r_{t+1}$ and next state $s_{t+1}$
            \STATE $T_t[s][a] \leftarrow T_t[s][a] + 1$
            \STATE Update $Q(s_t, a_t)$ using Q-learning algorithm
            % \STATE $Q(s_t, a_t) \leftarrow Q(s_t, a_t) + \alpha[r_{t+1} + \gamma Q(s_{t+1}, a_{t+1}) - Q(s_t, a_t)]$
            % \STATE $s_t \leftarrow s_{t+1}; a_t \leftarrow a_{t+1}$
            \STATE $s_t \leftarrow s_{t+1}$
        \UNTIL{$s_t$ is terminal or $max\_steps$} reached %(goal or aversive state)}
        \IF{$s$ is goal state}
            \FOR{each $s, a \in T_{list}$}
                \STATE $T_s[s][a] \leftarrow T_s[s][a] + 1$
            \ENDFOR
        \ENDIF
        \STATE \text{Compute $P_s \leftarrow T_s / T_t$}
        % \STATE \text{Compute $N_t$ for each $s \in T_{list}$ as $pos(s, T_{list}) +1$}
    \ENDFOR
\end{algorithmic}
\end{algorithm}

\section{Experimental scenario}
In the following section, we provide a description about the research experiments carried out. 
This includes details about probability, task hierarchies, and sub-tasks. 
In addition, we describe the rules of the training process.
In this work, a simple simulated scenario is used as this is an initial approximation for this approach in the context of hierarchical reinforcement learning scenarios\footnote{Code available at https://github.com/hugo12xx/memoria}.
Therefore, we aimed at obtaining preliminary results to analyze the method's feasibility in hierarchical contexts.

\subsection{Problem to resolve}
In this work, we focused on solving the spaceship escape problem. 
It consisted of an astronaut (agent) inside of a spaceship that needed to return home by crossing a wormhole. 
Prior to crossing the wormhole, the spaceship needed to first put up a shield before crossing the wormhole; otherwise, it would explode. 
Furthermore, various black holes existed in the environment. 
If the spaceship were to fall into one of these, it would immediately lose its course, and the spaceship would not be able to escape the black hole. 
Therefore, it would not be able to return home. 

\begin{figure}%[htbp]
\centerline{\includegraphics[width=\linewidth]{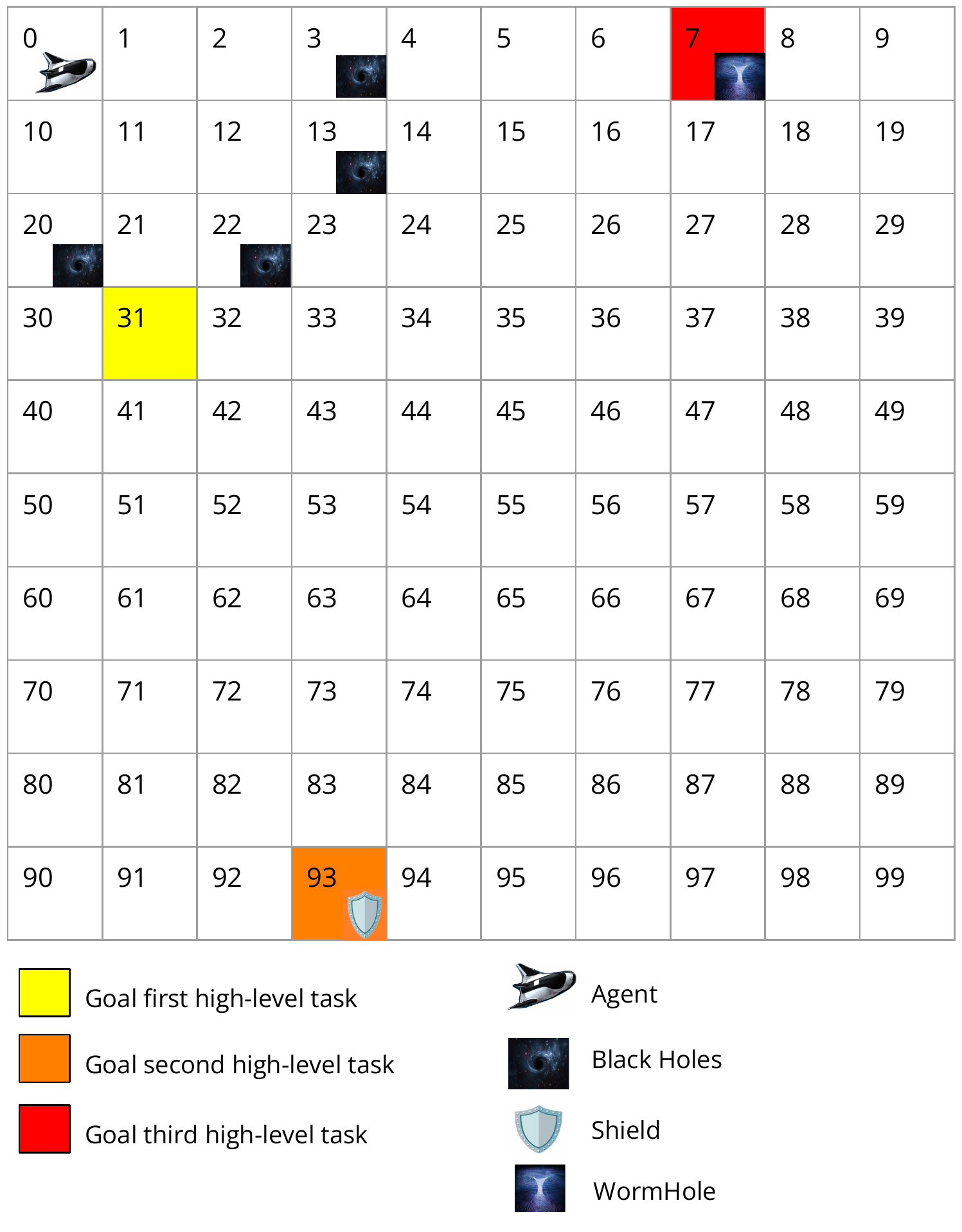}}
\caption{Possible states go from 0 to 99. 
State 0 (the place where the spaceship is located in the figure) represents the initial state. State 7 (wormhole) represents the final state (goal).
Black holes are present in states 3, 13, 20, and 22. 
It is not possible to return home if the spaceship falls into one of these black holes.
Furthermore, before going towards the wormhole to escape, the agent needs to pass through state 93 in order to collect the shield. 
Otherwise, if state 7 is reached without the shield, the objective will not have been achieved. }
\label{fig:estados}
\end{figure}

Figure~\ref{fig:estados} illustrates a map of the possible states including the black holes, the position of the shield, and the wormhole. 
Each location in the $10 \times 10$ maze grid is indexed with the number of states, as shown in this figure.
For example, the starting position for the spaceship, labeled as 0 is shown in the upper left hand corner of the maze. 
To return home, the position of the shield corresponds to state 93, and the position of the wormhole corresponds to state 7. 
It is important to highlight that if the spaceship reaches state 7 without first collecting the shield, then, the wormhole would have the same effect as a black hole. 
States 3, 13, 20, and 22 corresponded to the black holes. 
Therefore, initially, the spaceship was surrounded by black holes. 
As a result the ship needed to learn how to escape from this zone in order to later find the shield.

\subsection{Hierarchy of the tasks}
In order to solve the spaceship escape problem, we proposed using hierarchical reinforcement learning for the tasks that seeks to divide the problem into high-level actions. 
The defined actions are to reach state 31, then to reach state 93, and finally to reach state 7. 
As a result, the agent will learn how to solve this sequence of high-level tasks instead of learning all possible actions involved at the lower level. 
In this context, selecting any of the four displacement movements allowed for the spaceship corresponds to a low-level task, i.e., go up, down, left, or right.

As illustrated in Figure~\ref{fig:estados}, the spaceship starts from state 0 with the goal of reaching state 7 with the shield in order to escape and return home.
The details for all of the high-level tasks are shown below:

\begin{itemize}
    \item \textbf{First high-level task}: the first task the agent needed to perform was to escape from the upper left hand corner. 
    The black holes surrounded the spaceship in this corner. 
    Therefore, if it does not learn to escape this zone, it could not return home. 
    It was assumed that the agent will learn how to escape the zone with the black holes by reaching state 31, just below the zone of the black holes.
    
    \item \textbf{Second high-level task}: once outside the corner of the black holes, the agent can move at the lower/inferior part of the maze. 
    However, it may not go simply to state 7 to escape through the wormhole. 
    First, it must find the shield that was in state 93. 
    Thus, finding this state was the second high-level task.
    
    \item \textbf{Third high-level task}: once the spaceship has escaped from the black holes in the corner and collected the shield, the agent needed to solve the next high-level task, that was to find the wormhole that corresponded to exiting, or rather, reaching state 7 with the shield.
    
\end{itemize}

\subsection{Training rules}
The spaceship problem was represented by a Markov decision process and solved through reinforcement learning. 
Thus, some training rules were established. 
These are described in the following way:
\begin{itemize}
    \item The scenario allowed for the execution of four actions i.e., up, down, left, and right. 
    The agent may select any action that moved the spaceship by one box in the direction chosen.
    \item If the spaceship falls into a black hole, the training session should stop immediately, and the agent will receive a negative reward equal to a -100.
    \item The agent cannot exit the $ 10 \times 10$ grid. 
    The actions that would let it exit were limited.
    \item The agent should learn to escape from the zone of the black holes. 
    Once the agent is out of this zone, it receives a reward of 200.
    This state is reflected in Figure~\ref{fig:estados} with the color yellow and corresponds to state 31.
    \item In order to return home, the spaceship needed to collect the shield from state 93 (the color orange in Figure~\ref{fig:estados}). 
    After obtaining the shield, the agent received a reward of 200.
    \item Having collected the shield and reaching the goal state which is represented by the wormhole in state 7 (the color red in Figure 2), the agent received a reward equal to 500. 
    On the contrary, without the shield, the agent received a negative reward of -100, and the training episode ends.
    \item When the agent reached the goal or lost in some way, the training session ended, and a new one began.
    \item The maximum number of steps per episode for the first high-level task was $max\_steps = 10$, while for the second and third was $max\_steps = 100$.
\end{itemize}

\begin{figure}%[htbp]
\centerline{\includegraphics[width=0.95\linewidth]{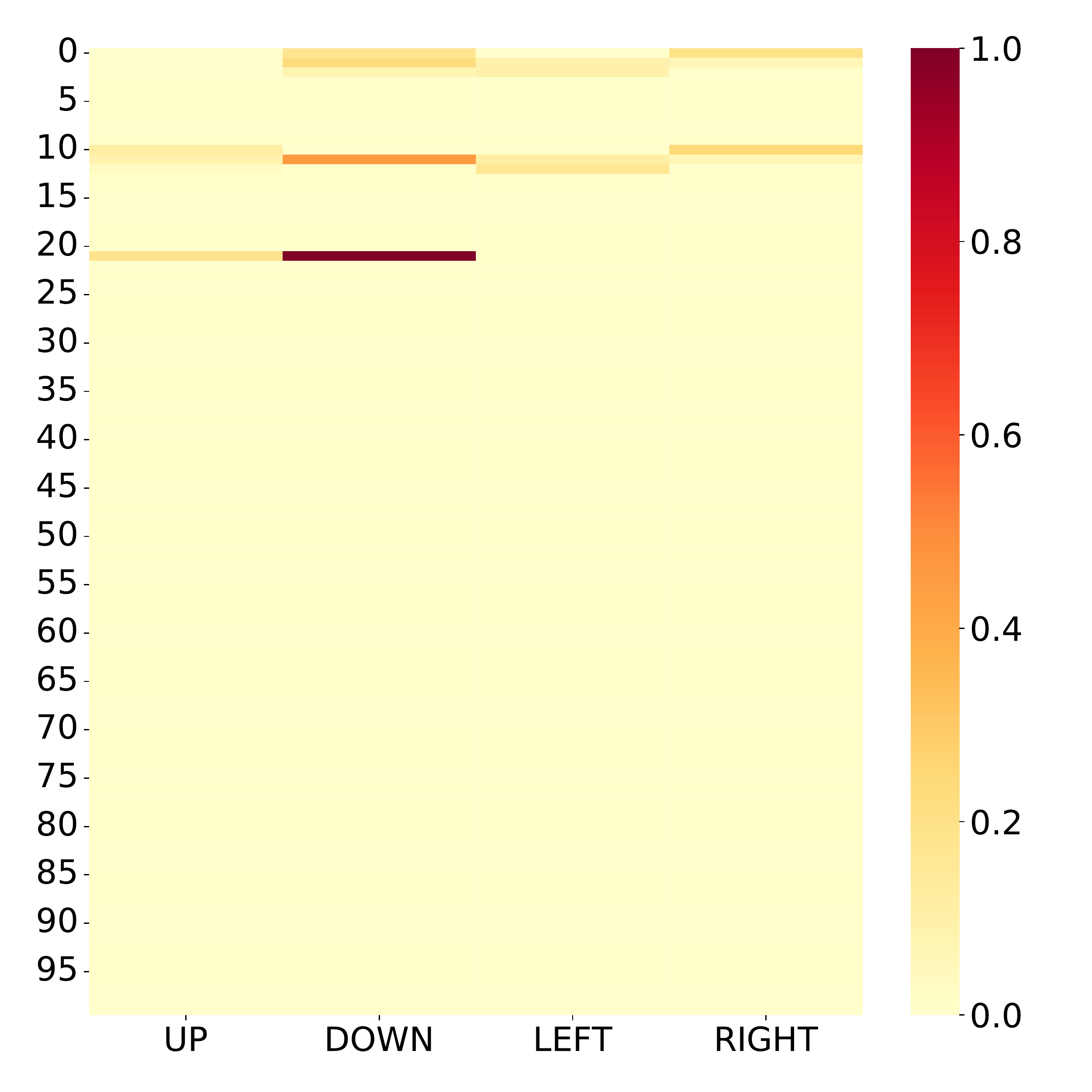}}
\caption{Heat map with the probabilities of success for the first high-level task. 
The Y axis represents the 100 possible states and the X axis shows the four defined actions. 
Higher probabilities of success are observed from state 21 as it is more possible to escape the black holes zone and reach the goal state.}
\label{fig:probprimeratarea}
\end{figure}

\section{Results}
In the following section, the results from the experiments are presented. 
For the agent, its position was located in the starting position corresponding to the high-level task that was to be learned.
For the experiments, the Q-learning algorithm was used with learning rate $\alpha=0.00001$, discount rate $\gamma = 0.9$ and the action selection method $\epsilon$-greedy with $\epsilon = 0.7$. 
%Therefore, the agent only selected the actions with the most possibilities to accomplish the objective 30\% of the times. 
%On the other hand, 70\% of the time, the agent will choose a random action to explore a possible new path. 
The training episodes varied for each high-level task, and they were determined empirically related to each task and to the training parameters.

Next, the probabilities of success are shown for each state-action pair and each high-level task. 
Additionally, the probabilities of success for the global task are also shown. 
Once the probabilities of success were calculated, these could be used together with a template to generate goal-driven explanations.

\begin{figure}%[htbp]
    \centerline{\includegraphics[width=0.95\linewidth]{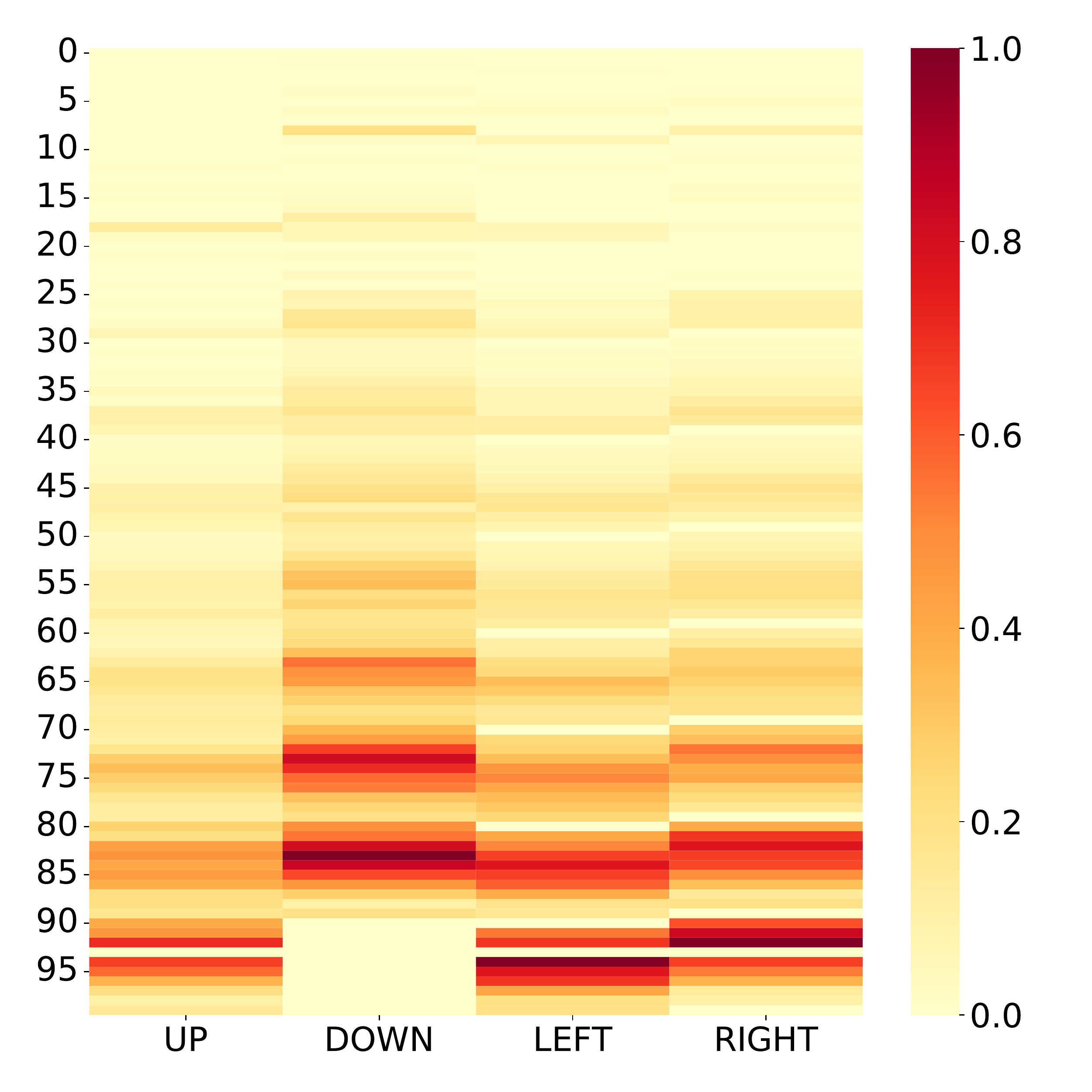}}
    \caption{Heat map of the probabilities of success for the second high-level task. 
             The Y axis represents the 100 possible states, and the X axis shows the four defined actions. 
             State-action pairs that move the agent closer to state 93 show a higher probability of successfully collecting the shield.}
    \label{fig:probsegundatarea}
\end{figure}

\subsection{Training for the first high-level task}
For the first high-level task, the agent began from state 0 as the starting point and state 31 as the goal. 
In this task, the agent only sought to escape the black holes zone without falling into them.
During the training, 10,000 episodes were used with 10 iterations in each one. 
In spite of the existence of different escape routes, the best paths for the agent to understand were states 0, 1, 11, 21, and 31.

Figure~\ref{fig:probprimeratarea} shows the probabilities computed during the first high-level task in a heat map.
As can be seen, states 3, 13, 20, and 22, or any action that made the agent enter any of these states always had possibility of success 0 since these states correspond to the black holes. 
In the same way, it was observed that the actions that got the agent closer to the goal presented the greater probability. 
For example, in state 21, the action go down showed 100\% probability of success since always when the agent chose to move down from state 21 would escape the black-holes zone and complete the first high-level task.
Similarly, on state 11 the action go down had a high probability of success but not of 100\%.
This happened because moving down from state 11 reached state 21 from which the agent could go left or right falling into a black hole.
The action go left or right from state 11 showed a low probability of success but was not null, as the agent would move away from the goal but still had a probability of escape.
%\todo{NOTA: Este parrafo estaba traducido en la subseccion de entrenamiento para la segunda tarea de alto nivel, por lo que hice mi propia traducción por que no lo encontraba. Lo dejo aca para decidir cual ocupar. The same for state 11. The descending action had a higher probability of success, but not 100\%. Descending from state 11, the agent reached box 21from which the spaceship could still fall into a black hole by moving to the left or right. The action to the right or to the left from box 11 would also show a probability of success lower but not null. Despite moving away from the goal, it still had the possibilities of escaping.}

Considering the above, when the agent is in state 11 and performs the action go down, it is possible to ask it: \textit{ why didn't you go left on the last action?.}
The agent with the probability of success as a basis could respond using a template in the following way:
\textit{I did not move to the left since carrying out this action, I would only have a 25\% probability of escaping the black holes while going down, I have a 60\% probability of escaping.}

\begin{figure}%[htbp]
    \centerline{\includegraphics[width=0.95\linewidth]{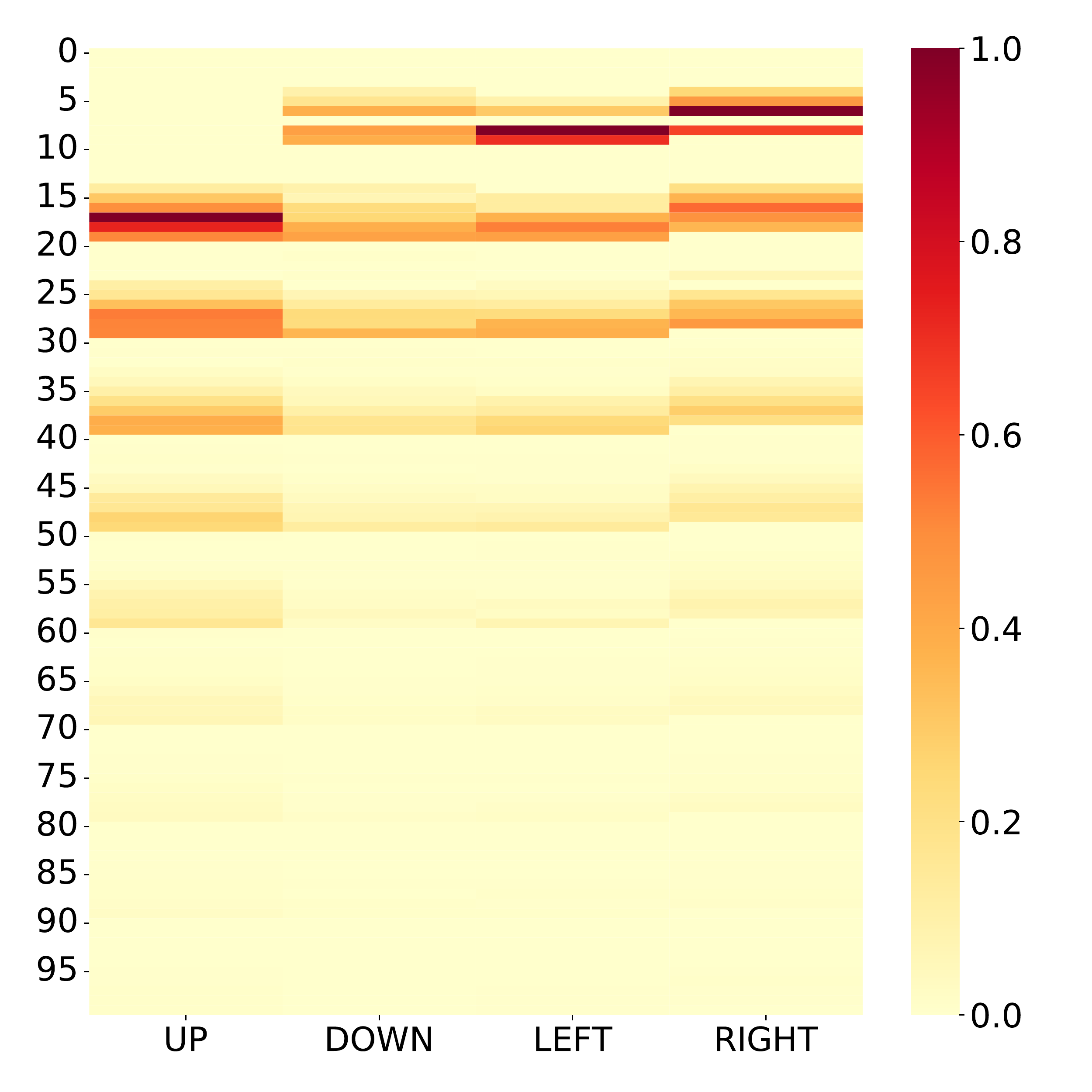}}
    \caption{Heat map of the probabilities of success for the third high-level task. 
    The Y axis represents the 100 possible states and the X axis shows the four defined actions.
    Actions that move the agent into state 7 show a higher probability of success as they allow reaching the wormhole and completing the mission.}
    \label{fig:probterceratarea}
\end{figure}

\subsection{Training for the second high-level task}
In the second high-level task, the agent had state 31 as the starting point and 93 as goal state where it sought to collect the shield. 
For this task 15,000 episodes were used with a maximum of 100 steps in each one. 
Of all of the possible paths, the best one the agent found included states 31, 41, 51, 61, 71, 81, 91, 92, and 93.

Figure~\ref{fig:probsegundatarea} illustrates the probabilities of success computed during the second high-level task. 
The closer the agent comes to the goal (state 93) the greater the probabilities of escaping the black holes. 
In this case, there are three actions that provided 100\% probability of success, that corresponds to the actions that directly move the spaceship to state 93.

% The sane for state 11. The descending action had a higher probability of success, but not 100\%. Descending from state 11, the agent reached box 21from which the spaceship could still fall into a black hole by moving to the left or right. The action to the right or to the left from box 11 would also show a probability of success lower but not null. Despite moving away from the goal, it still had the possibilities of escaping.

% Considering the previous information, when the agent finded in state 11 and carried out the downward action, it was possible that it could ask: Why did you not move to the left for the last action? As a basis, the agent used abilities that provided 100\% probability of success that corresponds to the actions that directly move the spaceship to state 93.
The three state-action pairs that moved to state 93 were from state 92 to right, from state 83 to down, and from state 94 to left. 
That is, these state-action pairs guarantee completing the mission of collecting the shield. 
% The possible action decisions from states are found in all forms of the goal. 
% Furthermore, they showed a high probability of success.
Other possible actions from those states also showed a high probability of success, because they are close to the goal.
For example, from state 83, moving down yields 100\% success while up, right, and left maintained a high probability of success, but it did not guarantee completing the mission.

Considering this scenario, when the agent finds itself in state 83, and carries out the action to move down, the user could ask: \textit{Why did you move down with the last action?}
Based on the probabilities of success computed, the agent using a template could respond: \textit{I moved down because in doing so, I have a 100\% probability of collecting the shield}.

\begin{figure}%[htbp]
    \centerline{\includegraphics[width=0.95\linewidth]{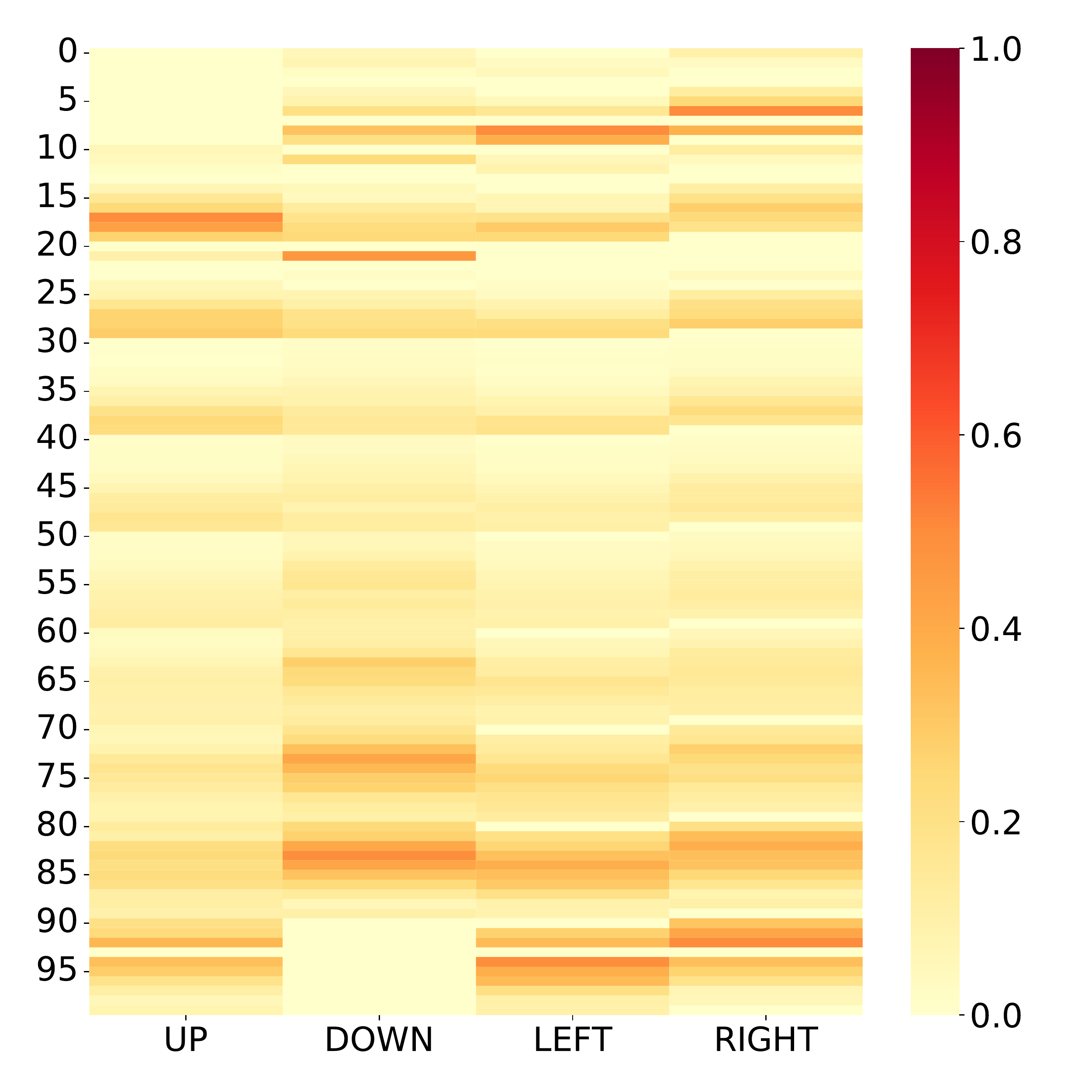}}
    \caption{Heat map with the global probabilities of success. 
    %The Y axis represents the 100 possible states and the X axis shows the four defined actions.
    The highest probabilities are observed as the different high-level tasks were completed. 
    However, no state-action pair obtained a probability of success of 100\% because when considering the global problem, no single action guarantees returning home. 
    }
    \label{fig:probgeneral}
\end{figure}

\subsection{Training for the third high-level task}

In the third high-level task, the training for the agent had as a starting point state 93 and as a goal state 7. 
In this task, the agent sought to cross the wormhole located in state 7 and return home. 
For this task, 20,000 episodes were used with 100 iterations in each one of the episodes. 
Of all of the paths explored, the best path found corresponds to the route between states 93, 94, 95, 96, 97, 87, 77, 67, 57, 47, 37, 27, 17, and 7. 
The probabilities of success computed during the third high-level task are illustrated in Figure~\ref{fig:probterceratarea}. 
As the agent got closer to the goal (state 7), the highest probabilities of success are observed. 
In this case, three state-action pairs provided 100\% probability of success.  
The three possibilities with which state 7 was entered are: from state 6 to the right, from state 17 up, and from state 8 to the left. 
When one of these actions was executed in these states, the agent would enter into the wormhole. 
That is, it would complete the mission, and the spaceship would return home.

Using the probabilities of success as a basis, the agent one more time can explain its behavior. 
For example, from state 16 and executing the action to move up, it was possible to ask the agent: \textit{Why did you not move to the left in the last action?} 
Then, the agent could respond using a template in the following way: \textit{I did not move to the left because doing so, I would have only have a 30\% probability of reaching the wormhole and of returning home, while moving up I have an 80\% probability of completing the mission successfully.}

\subsection{Global probability of success}

As demonstrated in the previous experimental results, the probability of success for each one of the high-level tasks was obtained. 
However, this does not allow the agent to show the probability of success for the problem in global terms, but it allows showing only one high-level task at a time. 
Therefore, we computed a matrix for the global probabilities of success by averaging the three individual matrices for the probabilities of success obtained previously. 
This matrix is illustrated in Figure~\ref{fig:probgeneral}. 
The matrix for the global probabilities of success, unlike the individual matrices, did not represent a particular high-level task, but it combined the three individual matrices together in order to obtain the general probabilities of success for the global
problem. 
Through the global probability, we can better understand the agent’s behavior. 
For example, from states 0, 1, 2, 3, 4, 5, 6, 7, 8 and 9, the action up, always had 0\% probability of success. 
From states 0, 10, 20, 30, 40, 50, 60, 70, 80 and 90, the action of moving the agent to the left also had a 0\% probability of success. 
In addition, from states 9, 19, 29, 39, 49, 59, 69, 79, 89 and 99, the action to the right, had 0\% probability of success as well.
And finally, from states 90, 91, 92, 93 94, 95, 96, 97, 98 and 99, the action down, had also 0\% probability of success as the agent was not allowed to exit the $ 10 \times 10 $ grid.
Moreover, it could be seen the behavior when entering the black holes. 
For example, any action that entered states 3, 13, 20, had a 0\% probability of success.

It is important to highlight that in the global matrix, unlike in the individual matrices, no action provided 100\% probability of success because indeed no state-action pair in this scenario guaranteed success. 
It is important to remember that upon entering the wormhole in state 7, it is not guaranteed that the spaceship could escape and return home. 
The escape only occurred if the shield had been collected beforehand. 
In this case, the success of the mission was reduced to the question: Upon entering the wormhole, does the spaceship have the shield? 
If the spaceship does not had the shield, then, the effect would be the same as that of a black hole. 
With respect to the global probabilities of success, the same occurred with the actions that made the agent enter state 93 or collect the shield.
In the global matrix, these actions did not have 100\% probability of success since collecting the shield still did not guarantee escape. 
To successfully escape, the spaceship still needed to cross the wormhole and, along the way, it had the possibility of falling into a black hole. 
Overall, although the highest probabilities of success were obtained by completing the different high-level tasks, none reached 100\% success, as pointed out above, as no action could guarantee escape.

\section{Conclusions}
In this work, we have demonstrated that with the memory-based explainable reinforcement learning method using hierarchical training, it is possible for a learning agent to learn how to escape a scenario while computing probabilities of success to be used for explanations. 
Being a hierarchical scenario, the training was divided in order to complete high-level tasks. 
The explainability method was used to compute the global and individual matrices with the probability of success. 
Afterward, these were used as a basis for exploring agent's behavior using explainability. 
A matrix was used to explain high-level tasks (one matrix was created for each task). 
However, this only resulted in an explanation of how to complete an individual high-level task. 
To observe the agent's overall behavior, a global matrix was computed with the general probabilities by averaging the matrices obtained for each high-level task. 
The global matrix showed coherent probabilities of success for the agent when executing an action starting from a specific state being, therefore, a good basis to be used for generating explanations that can be understood by non-expert users.
Future work includes extending our research into more complex environments, especially continuous, in order to explore other explainability algorithms, such as the learning-based or introspection-based methods for hierarchical tasks.

\section*{ACKNOWLEDGMENT}
This research was partially financed by Universidad Central de Chile under the research project CIP2020013, the Coordenação de Aperfeiçoamento de Pessoal de Nível Superior—Brasil (CAPES)—Finance Code 001, Fundação de Amparo a Ciência e Tecnologia do Estado de Pernambuco (FACEPE), and Conselho Nacional de Desenvolvimento Científico e Tecnológico (CNPq)—Brazilian research agencies.

\bibliographystyle{IEEEtran}
\balance
\bibliography{bibliografia}

\end{document}